\documentclass[10pt,twocolumn,letterpaper]{article}

\usepackage{cvpr}
\usepackage{times}
\usepackage{epsfig}
\usepackage{graphicx}
\usepackage{amsmath}
\usepackage{amssymb}

\usepackage{booktabs}
\usepackage[ruled,vlined]{algorithm2e}
\newcommand{\dtoprule}{\specialrule{1pt}{0pt}{0.4pt}%
            \specialrule{0.3pt}{0pt}{\belowrulesep}%
            }

\cvprfinalcopy 

\def\cvprPaperID{000} 

\ifcvprfinal\fi

\begin{document}

\title{Cross-Domain Few-Shot Learning with Meta Fine-Tuning}

\author{John Cai \\
Princeton University \\
  {\tt jjcai@princeton.edu} 
\and 
Shen Sheng Mei \\
Pensees Pte Ltd \\
  {\tt jane.shen@pensees.ai}
}

\maketitle

\begin{abstract}
   In this paper, we tackle the new Cross-Domain Few-Shot Learning benchmark proposed by the CVPR 2020 Challenge. To this end, we build upon state-of-the-art methods in domain adaptation and few-shot learning to create a system that can be trained to perform both tasks. Inspired by the need to create models designed to be fine-tuned, we explore the integration of transfer-learning (fine-tuning) with meta-learning algorithms, to train a network that has specific layers that are designed to be adapted at a later fine-tuning stage. To do so, we modify the episodic training process to include a first-order MAML-based meta-learning algorithm, and use a Graph Neural Network model as the subsequent meta-learning module to compare the feature vectors. We find that our proposed method helps to boost accuracy significantly, especially when coupled with data augmentation. In our final results, we combine the novel method with the baseline method in a simple ensemble, and achieve an average accuracy of 73.78\% on the benchmark. This is a 6.51\% improvement over existing benchmarks that were trained solely on miniImagenet.
\end{abstract}

\section{Introduction}

In the past decade, vast improvements to visual recognition systems have been achieved by training deep neural networks on ever-expanding training data-sets. Often, the ability of these neural network models to generalize directly depends on the size and variance of the training data-set. Unfortunately, acquiring a large training data-set is costly due to the need for human annotation. Furthermore, when dealing with rare examples in medical images (e.g. rare diseases) or satellite images (e.g. oil spills), the ability to obtain labelled samples is limited. Moreover, there is vast scope for improvement, as the human visual system is far less data-hungry than current deep learning methods.

To address the limitations of traditional deep learning, few-shot learning methods have emerged to train models that predict new classes by seeing only a few labelled images per class. These methods have shown promising improvements over the last 5 years.

However, existing few-shot learning methods have been developed with the assumption that the training and test data-set arise from the same distribution. Domain shift would thus be an additional problem as it may prevent the robust transfer of features. 

Hence, the CVPR 2020 challenge has introduced a new benchmark that aims to test for generalization ability across a range of vastly different domains, with domains from natural and medical images, domains without perspective, and domains without color \cite{guo2019new}. This robust evaluation framework allows one to truly test for how few-shot learning models do when faced with sharp domain shifts. This is in contrast to previous attempts at considering domain-adaptation \cite{chen2019closer}, as the training and test domains previously used were still fairly close to each other.

The main contribution of this paper is the integration of fine-tuning into the episodic training process by exploiting a first-order MAML-based meta-learning algorithm (henceforth ``Meta Fine-Tuning''). This is done so that the network learns a set of initial weights that be easily fine-tuned on the support-set of the test domain. 

Second, this paper integrates the above Meta Fine-Tuning algorithm into a Graph Neural Network that exploits the non-Euclidean structure of the relation between the support set and the query samples.

Third, as the baseline code-base only implements data augmentation for the training process, we implement data augmentation on the support set during fine-tuning, and achieve a further improvement in accuracy.

Finally, we combine the above method with a modified fine-tuning baseline method, and combine them into an ensemble to jointly make predictions.

\section{Relevant Work}

The key method that this paper will build on is Graph Neural Networks. Graph-based convolutions can create more flexible representations of data beyond a simple Euclidean space \cite{bronstein2017geometric}. For the context of few-shot learning, after obtaining image features through deep learning networks, the problem can be reconstrued as a belief propagation problem. Under this framework, labels from labelled support examples are propagated to unlabelled query examples. Hence, by representing the set of support examples as a densely-connected undirected graph, we can add each query example in, and learn edge weights \cite{garcia2017few}.

Model-Agnostic Meta-Learning (MAML) was developed to train a network to learn a set of internal representations that can easily be adapted \cite{finn2017model}. It has been shown that the first-order approximations of the MAML algorithms, such as Reptile \cite{nichol2018first}, that ignore second-order derivatives perform as well on established benchmarks.

A key method for domain adaptation is to fix earlier feature layers, and fine-tune later feature layers on the support examples \cite{guo2019new}. This could help transfer high-level features, while retraining domain-specific features. 

Averaging the prediction scores of different models to reduce variance has been well-documented to achieve higher accuracy \cite{dvornik2019diversity}, which is why this paper explores the combination of models with different architectures.

\section{Methodology}
\subsection{Graph Neural Networks}
The Meta-Learning module we use is the Graph Neural Network, which has been used for cross-domain few-shot learning \cite{tseng2020cross}. To begin, I use a linear layer to project the feature vectors of dimensional \(F\) onto a lower-dimensional space \(d_k\) . Then, the GNN takes in the input signal \(S \in \mathbb{R}^{N_s + 1} \) where \(N_s\) is the number of support samples, with 1 vertex for the query sample. Then, a graph convolution layer \(GC(.)\) is performed with linear operations on local signals. This produces an output \(X^{(k+1)} = GC(X^{(k)}) \), with the GC layer having \(d_k \times d_{k+1}\) parameters. To learn edge features, a MLP takes in the absolute difference between the the output vectors of vertices in the graph. For more details, refer to the GNN few-shot learning paper \cite{garcia2017few}. 

\subsection{Meta Fine-Tuning}

The core idea of meta-fine tuning is that instead of fine-tuning a pre-trained model that was not trained explicitly for fine-tuning, we can use meta-learning to find a set of weight initializations that are intended to be fine-tuned. To this end, we apply and adapt the first-order MAML algorithm \cite{nichol2018first} and simulate the episodic training process. A first-order MAML algorithm can achieve comparable results with the second-order algorithm at a lower computational cost.

\begin{algorithm}[h]
\SetAlgoLined
 Initialize weights \(\phi_f\) for feature extractor and \(\phi_m\) for metric-learning module; \\ 
 \For{each episode}{
  Sample \(N_s\) support samples and \(N_q\) query samples \\  
  Freeze first \(L-k\) layers of feature extractor \\ 
  \For{step = 1,2,...,\(S\)}{
    Sample batch \(b\) from the \(N_s\) support samples \\
    Compute loss \(L_s\) on these support samples using linear classifier \\
    Update the last \(k\) layers of feature extractor using SGD or Adam
  }
  Obtain \(\Tilde{\phi}_{f(k)}\) = \(U_b^S(\phi)\), the updated weights for last \(k\) layers \\ 
  Combine \(\phi_{f(L-k)}\) and \(\Tilde{\phi}_{f(k)}\) to obtain new feature extractor \(\Tilde{\phi}_f\)\\
  Feed images through feature extractor and then through the metric-learning module \\ 
  Compute the loss \(L(\Tilde{\phi}_f, \phi_m)\) on the \(N_q\) query samples and compute update \(g_{f(L-k)}\), \(g_{f(k)}\), \(g_m\) for all model parameters using Adam \\
  Update initial parameters using learning rate \(\theta \): \\ 
  \(\phi_{f(L-k)} = \phi_{f(L-k)} - \theta g_{f(L-k)}\) \\
  \(\phi_{f(k)} = \phi_{f(k)} - \theta g_{f(k)}\) \\
  \(\phi_m = \phi_m - \theta g_m\) \\
 }
 \caption{Meta Fine-Tuning Algorithm}
\end{algorithm}

The algorithm is model-agnostic and can be used with any existing metric-learning module.  However, typically a metric-learning module with weights should be used, as a completely non-parametric module like Prototypical Networks, which uses nearest centroid, may not be able to compare subsequent fine-tuned image features in a robust way. In this paper, the GNN described above is used because it is a highly trainable and flexible meta-learning module, and hence it can learn how to compare features that have been fine-tuned when trained on the fine-tuning process.

The method can also be applied to a model of any backbone depth, and you can freeze up to any number of layers. For this paper, we freeze the last ResNet block in ResNet10.

For further visualization, we include the figure below.

\begin{figure}[h]
\begin{center}
   \includegraphics[width=0.99\linewidth]{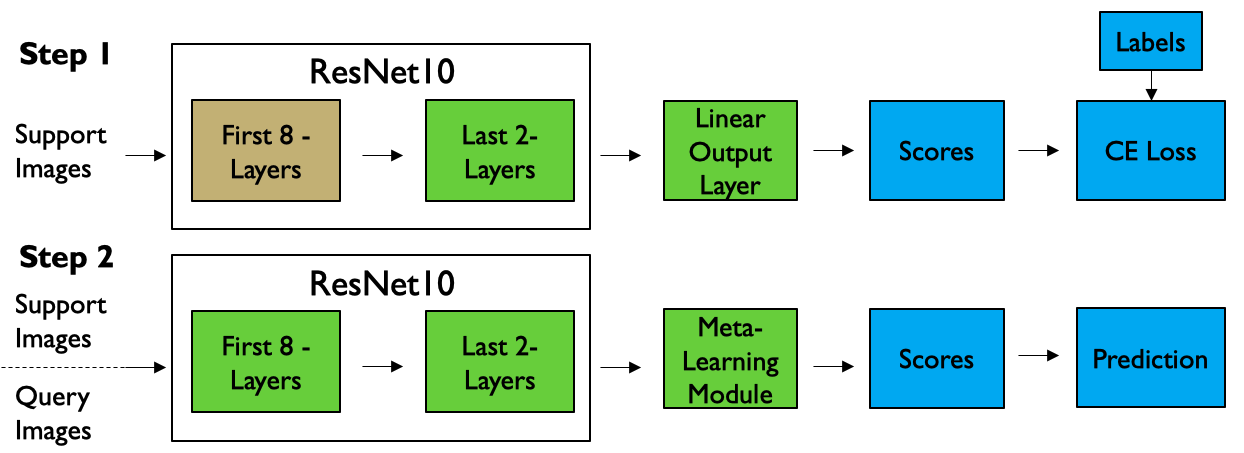}
\end{center}
   \caption{Meta-learning with fine-tuning. Green are trainable, brown are frozen. At test time, all will be frozen in step 2. }
\label{fig:long}
\label{fig:onecol}
\end{figure}

\begin{table*}[t]
  \centering
  \begin{tabular}{@{}ccccc@{}}
    \dtoprule
    No. of Shots  & CropDisease    & EuroSAT & ISIC & ChestX  \\
    \midrule
    
    5 & 96.27\% \(\pm\) 0.40\% &	89.83\% \(\pm\) 0.46\% &	61.71\% \(\pm\) 0.44\% &	28.44\% \(\pm\) 0.43\% \\ 
    20 & 98.91\% \(\pm\) 0.19\%	& 93.90\% \(\pm\) 0.34\%	& 65.29\% \(\pm\) 0.56\%	& 35.62\% \(\pm\) 0.47\% \\ 
    50 & 99.48\% \(\pm\) 0.13\%	& 96.08\% \(\pm\) 0.25\% & 75.13\% \(\pm\) 0.56\% &	44.70\% \(\pm\) 0.64\% \\ 
    \bottomrule
  \end{tabular}
  
  \caption{Final Proposed Model: Meta Fine-Tuning GNN + Modified Baseline Fine-Tuning + Data Augmentation}
\end{table*}

\begin{table*}[t]
  \centering
  \begin{tabular}{@{}ccccc@{}}
    \dtoprule
    No. of Shots  & CropDisease    & EuroSAT & ISIC & ChestX  \\
    \midrule
    5 & 88.72\% \(\pm\) 0.53\% &	80.45\% \(\pm\) 0.54\% &	47.20\% \(\pm\) 0.45\% & 25.96\% \(\pm\) 0.46\%	 \\ 
    20 & 95.76\% \(\pm\) 0.65\%	& 87.67\% \(\pm\) 0.44\%	& 59.95\% \(\pm\) 0.45\%	& 31.63\% \(\pm\) 0.49\% \\ 
    50 & 97.87\% \(\pm\) 0.48\%	& 90.93\% \(\pm\) 0.45\% & 65.04\% \(\pm\) 0.47\% &	37.03\% \(\pm\) 0.50\% \\ 
    \bottomrule
  \end{tabular}
  
  \caption{Previous Benchmark's Best Model ``Ft-Last1" trained on MiniImagenet from \cite{guo2019new}}
\end{table*}

During step 1 (Meta Fine-Tuning), only support examples are used, and the first 8-layers are frozen. A linear classifier on the ResNet10 features is used to predict the support labels, and the last 2-layers are updated accordingly using CE Loss for 5 epochs. At step 2, all layers are updated using the episodic training loss.  At prediction stage on the test domain, all layers in the ResNet10 are frozen in step 2.

\subsection{Data Augmentation}
For data augmentation during training, we stick to the default parameters in the codebase. For data augmentation during testing, we sample 17 additional images from the support images (which we know the labels of), and perform jitter, random crops, and horizontal flips (if applicable) on a randomized basis. In the fine-tuning process, we weight the original images more by exposing the model to the original images more frequently. At the final prediction stage, only base images (which are center-crops) are used for both support and query images.

\subsection{Combining Scores in the Ensemble}
The baseline fine-tuning model used in the ensemble was modified so that we only fine-tune then last ResNet Block, and use an Adam optimizer with weight decay over 20 epochs rather than the default Stochastic Gradient Descent. 

For our final submission results, we combine the predictions from the modified baseline fine-tuning model and the meta fine-tuning GNN model by normalizing the scores using a softmax function so that the scores from each model sum to 1 and are between 0 and 1, which ensures that each model is given equal weight in the prediction.  Then we add them together and take argmax. I also implement transduction, as it is allowed in Track 1 of the competition.

\subsection{Memory Requirements of GNN on 50-shot}
The GNN builds a densely-connected graph between all the support samples and each new query sample. Hence, the space requirement for the 50-shot is tremendous, as the memory requirements scale up at \(O(n^2)\). Thus, in order to fit the model onto a 16GB Tesla V100, we average every 2 support samples' feature vectors into 1, so that we obtain 25 nodes for the GNN. This approximation scheme can be extended for higher number of shots (\(N_s>50\)) as well.

\section{Submission Results (Table 1)}

\subsection{Experimental Setup}
The experimental setup involves training on miniImagenet and testing on CropDisease, EuroSAT, ISIC and ChestX. Models are trained for 400 epochs and then meta fine-tuned for 200 epochs. During training, we augment the training dataset using image transformations, following the protocol in \cite{chen2019closer}. We modify the evaluation code in \cite{guo2019new} to ensure that the same test images are used and add checks to ensure that the same base images are used per-episode. The code can be found at \url{https://github.com/johncai117/Meta-Fine-Tuning}.

\subsection{Proposed Model}

As shown on Table 1 and Table 2, the final proposed model vastly outperforms the previous benchmark introduced by \cite{guo2019new}. The average accuracy across all 12 tasks in the proposed model (Table 1) is 73.78\% while the average accuracy in the previous benchmark (Table 2), which is at 67.27\%. This is a 6.51\% improvement in the benchmark model that was trained solely on the miniImagenet dataset.

Even if we compare the accuracy with the performance of the benchmark model that was trained on multiple datasets (while the proposed model is not, due to the requirements of the challenge), the proposed model still has a vast improvement, as the previous IMS-f model achieved an average accuracy of 68.69\% \cite{guo2019new}.

We can also further observe that the improvement in accuracy is most pronounced at the 5-shot level, with a 8.48\% improvement in accuracy over the baseline Ft-Last1 model. This is followed by a 6.38\% improvement at the 50-shot level and a 4.68\% improvement at the 20-shot level. The non-linear improvement in the model may be attributed to the effect of data augmentation versus meta fine-tuning: data augmentation likely has the most effect when the number of support examples is very low, while fine-tuning has the most effect when the number of support examples is very high. This is supported by subsequent analysis.

\begin{table*}[t]
  \centering
  \begin{tabular}{@{}ccccc@{}}
    \dtoprule
    No. of Shots  & CropDisease    & EuroSAT & ISIC & ChestX  \\
    \midrule
    5 & 96.14\% \(\pm\) 0.43\% & 87.13\% \(\pm\) 0.58\% & 53.00\% \(\pm\) 0.45\%	& 26.76\% \(\pm\) 0.45\% \\ 
    20 & 98.66\% \(\pm\) 0.43\%	& 95.01\% \(\pm\) 0.33\% & 62.72\% \(\pm\) 0.73\% & 32.83\% \(\pm\) 0.45\% \\
    \bottomrule
  \end{tabular}
  
  \caption{Single Model Study: Meta Fine-Tuning GNN + Data Augmentation}
\end{table*}

\begin{table*}[t]
  \centering
  \begin{tabular}{@{}ccccc@{}}
    \dtoprule
    No. of Shots  & CropDisease    & EuroSAT & ISIC & ChestX  \\
    \midrule
    5 &  92.23\% \(\pm\) 0.46\% & 82.67\% \(\pm\) 0.50\% & 61.76\% \(\pm\) 0.50\% & 31.60\% \(\pm\) 0.41\% \\
    20 & 95.95\% \(\pm\) 0.30\% & 87.84\% \(\pm\) 0.46\% & 60.32\% \(\pm\) 0.59\% & 35.91\% \(\pm\) 0.42\% \\ 
    \bottomrule
  \end{tabular}
  
  \caption{Single Model Study: Modified Baseline Fine-Tuning + Data Augmentation}
\end{table*}

\section{Further Analysis of Results}

\subsection{Single Model Study}
For brevity, we only show results for 5 and 20 shot for the individual models. We see a clear pattern that meta fine-tuning is contributing more to improve accuracies on domains that are close to the training domain. We also see that baseline fine-tuning + DA is most effective at domains that are more distant from the training domain such as ChestX. We also find that the improvement going from 5 to 20 shots is less pronounced for baseline fine-tuning + DA.

\subsection{Ablation Study: GNN and Simple Fine-Tuning}

In this ablation study, we investigate which parts of the system are delivering superior performance. Simple fine-tuning refers to taking an existing GNN model and simply fine-tuning the last ResNet block. For brevity, we present results for EuroSAT and ISIC below for 20-shot.

\begin{table}[h]
\begin{center}
\begin{tabular}{|l|c|c|}
\hline
GNN Method & EuroSAT & ISIC\\ 
\hline\hline
No FT & 86.57\% \(\pm\) 0.63\% & 52.32\% \(\pm\) 0.64\% \\
Simp FT & 90.30\% \(\pm\) 0.47\% & 61.04\% \(\pm\) 0.64\% \\
Simp FT + DA & 94.60\% \(\pm\) 0.37\% & 63.34\% \(\pm\) 0.72\% \\
Meta FT + DA & 95.01\% \(\pm\) 0.33\% & 62.72\% \(\pm\) 0.73\% \\
\hline
\end{tabular}
\end{center}
\caption{20-Shot: Further experiments with fine-tuning of GNN. DA refers to data augmentation}
\end{table}

From above, we see that Meta Fine-Tuning and Simple Fine-Tuning achieve comparable results when paired with domain adaptation. It can be seen that in domains that are more similar to miniImagenet, Meta Fine-Tuning performs slightly better, while for domains further away from the training set, Simple Fine-Tuning performs better.

\section{Conclusion}

In this paper, we have developed a model that outperforms the benchmark by 6.51\%. This was done by extending first-order MAML algorithms to Meta Fine-Tuning, and combining this with GNN, data augmentation and ensemble methods . Further analysis suggests that Meta Fine-Tuning does especially well at domains close to the source domain while simple fine-tuning with data augmentation works better on domains that are further away. One reason is that learning how to fine-tune on miniImagenet may have made the model less optimized for fine-tuning on distant domains.

Still, most practical applications for cross-domain few-shot learning would not involve such a radical domain-shift, given that there exists a set of diverse datasets that may be closer to each domain than miniImagenet. Hence, Meta Fine-Tuning would still have significant utility. Further research can be done to train a meta fine-tuning model that is more domain-agnostic.

{\small
\bibliographystyle{ieee_fullname}
\bibliography{egbib}
}

\end{document}